\documentclass{article}
\usepackage{spconf,amsmath,graphicx}
\usepackage{amsmath,amsfonts,bm}
\usepackage{algorithm,algorithmicx}
\usepackage{algpseudocode}  
\usepackage{multirow}
\usepackage{booktabs}
\usepackage{color}

\DeclareMathOperator*{\minimize}{\text{minimize}}
\DeclareMathOperator*{\maximize}{\text{maximize}}
\DeclareMathOperator*{\st}{\text{subject to}}
\DeclareMathAlphabet\mathbfcal{OMS}{cmsy}{b}{n}

\algnewcommand{\LineComment}[1]{\State \(\triangleright\) #1}

\title{
Towards an Efficient and General Framework of Robust Training \\for Graph Neural Networks
}
\name{Kaidi Xu$^1$,  Sijia Liu$^2$,  Pin-Yu Chen$^2$, Mengshu Sun$^1$, Caiwen Ding$^3$, Bhavya Kailkhura$^4$, Xue Lin$^1$
}
\address{
$^1$Electrical \& Computer Engineering, Northeastern University, Boston, USA\\
$^2$MIT-IBM Watson AI Lab, IBM Research\\
$^3$Computer Science and Engineering, University of Connecticut, Storrs, USA\\
$^4$Lawrence Livermore National Laboratory, Livermore, USA
}

\begin{document}
%
\maketitle
\begin{abstract}

Graph Neural Networks (GNNs) have made significant advances on several fundamental inference tasks. As a result, there is a surge of interest in using these models for making potentially important decisions in high-regret applications. However, despite GNNs' impressive performance, it has been observed that carefully crafted perturbations on graph structures (or nodes attributes) lead them to make wrong predictions. Presence of these adversarial examples raises serious security concerns. Most of the existing robust GNN design/training methods are only applicable to white-box settings where model parameters are known and gradient based methods can be used by performing convex relaxation of the discrete graph domain. More importantly, these methods are not efficient and scalable which make them infeasible in time sensitive tasks and massive graph datasets. To overcome these limitations, we propose a general framework which leverages  the greedy  search algorithms and zeroth-order methods to obtain robust GNNs in a generic and an efficient manner. On several applications, we show that the proposed techniques are significantly less computationally expensive and, in some cases, more robust than the state-of-the-art methods making them suitable to large-scale problems which were out of the reach of traditional robust training methods.

\let\thefootnote\relax\footnotetext{This work was performed under the auspices of the U.S. Department of Energy by Lawrence Livermore National Laboratory under Contract DE-AC52-07NA27344.}

\end{abstract}
\section{Introduction}
\label{sec:intro}

Graph structured representations are one of the most commonly encountered data structure that naturally arises in nearly all scientific and engineering application~\cite{gama2018convolutional}. With the widespread use of networks  neuroscience,  molecular chemistry and other fields, it is not surprising that machine learning on graph data has become a key learning tool. Specifically, Graph neural networks (GNNs) have made significant advances on several fundamental tasks ranging from node classification to graph classification~\cite{kipf2016semi,xu2018how}. Despite GNNs’ impressive performance on inferring from graph data, their susceptibility to test-time adversarial examples (i.e., carefully crafted perturbations to fool these models) is a major hurdle in a universal acceptance of GNN solutions in several high-regret applications. These results are consistent with the adversarial attacks on images \cite{liu2018zeroth,hogan2018universal,zhao2019design,xu2019interpreting,xu2018structured,chen2019zo,ye2019adversarial,zhao2019admm}.

Recently, a few attempts have been carried out in the robust machine learning community to devise robust GNN training methods, e.g., adversarial training. 
Existing robust training methods for graph data \cite{zugner2019certifiable,xu2019topology} utilize convex relaxation to make the search domain continuous so that first-order methods can be used. Unfortunately, these approaches are highly inefficient (or time consuming) due to additional computations involved in the relaxation procedure and are infeasible to large-scale graphs. Furthermore, in practice the gradient information may not even be available, e.g., either due to inaccessible of the model parameters or model itself is discrete and non-continuous. In such scenarios, existing robust training methods fall short. To overcome these drawbacks, this paper proposes a general framework which leverages the greedy search algorithms and zeroth-order methods to obtain robust GNNs. Specifically, to address the inefficiency/scalability issue, we propose Greedy Topology Attacking method so that gradient-based adversarial training becomes plausible for massive size graphs. Next, to address the unavailability of gradient information issue, we propose Zeroth-order based Greedy Attack for gradient-free adversarial training. 
Benchmarking on node classification tasks using GNNs, our Greedy Topology Attack method can achieve similar performance with current state-of-the-art attacks with significant speed up subject to the same topology perturbation budget. This demonstrates the effectiveness of our attack generation method through the lens of greedy search algorithm. Next, by leveraging our proposed  greedy topology attack, we benchmark the robustness of our adversarial training technique for GNNs under different attacks and applications. We  show  that  the  proposed  adversarial training technique is significantly less computationally  expensive and, in some cases,  more  robust  than  the  state-of-the-art methods making them suitable to large-scale problems.

\section{Related Works}
 Some recent attentions have been paid to the vulnerability of GNNs. The authors in \cite{dai2018adversarial} investigated test-time non-targeted adversarial attacks on both graph classification and node classification problem by adding or deleting edges from a graph. The authors in \cite{zugner2018adversarial} considered both test-time (evasion) and training-time (data poisoning) attacks on node classification task. Different with~\cite{dai2018adversarial}, the node attributes  were allowed to modify in~\cite{zugner2018adversarial}. Their  algorithm is for targeted attacks on single node. It was shown that small perturbations on the graph structure and node features are able to misclassify the target node. 
 In \cite{zugner2019adversarial},  training-time attacks on GNNs were also studied for node classification by perturbing the graph structure. The authors solved a min-max problem in training-time attacks using meta-gradients and treated the graph topology as a hyper-parameter to optimize. 

On the defense side, very recent work~\cite{xu2019topology} proposed adversarial training on GNNs, however, this method needs to perform convex relaxation, probabilistic sampling and bisection which pay extra computational cost and is not feasible to large graph datasets. Different from the adversarial training, \cite{zugner2019certifiable} considered the problem of certifiable robustness and also achieved robust model under a pre-specific bound.

\section{Problem Statement}
\label{sec: GNN}
A GNN predicts the class of an unlabeled node under the graph topology.
The $k$-th layer computation in a GNN can be formulated as
\begin{align}
        \ H^{(k)} = \sigma \left (  \tilde{ A}  H^{(k-1)} (  W^{(k-1)} )
        \right ),
\end{align}%
where $H^{(1)} = X$ is the node feature matrix.
Here $\tilde { A}$ is defined as a normalized adjacency matrix $\tilde{ A} = \hat{ D}^{-1/2} \hat{ A} \hat{ D}^{-1/2}$, where $\hat{ A} =  A +  I$ and $A \in \{0,1\}^{N\times N}$ refers to the adjacency matrix.
We refer readers to \cite{kipf2016semi} for more details about GNN.
$\sigma$ is the ReLU activation function and $W$ is weight matrices.

\subsection{Adversarial Training for Graphs}
Following \cite{madry2018towards}, we consider an adversarial variant of standard Empirical Risk Minimization (ERM), where we minimize the risk over adversarial examples:
\begin{align}\label{eq:  robust_train}
    \begin{array}{cc}
\displaystyle \minimize_{W}   \maximize_{ A^\prime \in \mathcal{C}} \,  \mathcal{L}(A^\prime, W; X, y_L).
    \end{array}
\end{align}
The training loss $\mathcal{L}$ is cross-entropy error over all training data $(X, y_L)$. $A^\prime$ is the perturbed adjacency matrix by the adversarial attack.
Here  $\mathcal{C}$ are the constrains such as the maximum number of edges that can be perturbed, $A^\prime$ should be symmetric and most importantly, $A^\prime \in \{0,1\}^{N\times N}$.

Intuitively, adversarial training injects adversarial examples into training data via inner maximization problem to increase robustness of outer risk minimization (or training) problem.
The performance of adversarial training heavily depends on (a) the quality of the adversarial examples, and (b) the efficiency/scalability of the attack algorithms. Note that discrete constraint $\mathcal{C}$ in problem~\eqref{eq:  attack_loss} make the gradient descent inapplicable. As mentioned before, current approaches to handle these constraints are computationally inefficient in solving inner optimization problem. Therefore, we first propose efficient algorithms to generate adversarial attacks via greedy algorithms.

\subsection{Efficient Adversarial Example Generation}
In this section, we propose two approaches to generate adversarial examples based on greedy algorithms and zeroth-order methods.

\subsubsection{Greedy Topology Attack} \label{sec:attack}

We fix node attributes $X$ and only consider edge perturbations as the adversarial attack. We find the perturbed adjacency matrix $A^\prime$ to minimize the negative of training loss
\begin{align}\label{eq:  attack_loss}
    \begin{array}{ll}
\displaystyle \minimize_{A^\prime} & -\mathcal{L}(A^\prime ; W, X, y_L)\\
      \st    & A^\prime \in \mathcal{C}.
    \end{array}
\end{align}
 Next, we introduce our Greedy Topology Attack (GTA) in Algorithm~\ref{alg: greedy_attack} which is able to handle discrete constraints.

\begin{algorithm}
\caption{Greedy topology attack (GTA)}
\begin{algorithmic}[1]
\State \textbf{Input:} graph $G = (A,X)$, number of maximum edges can be changed $M$, greedy search step $n$, label $y_L$
\State \textbf{Output:} Modified adversarial adjacency matrix $A^\prime$
\State $A^\prime \leftarrow A$\;
\State $R \leftarrow \emptyset$
\While{{$\|A^\prime - A\|_0 < 2M$}}
\LineComment{Compute gradient of Eq.~\eqref{eq:  attack_loss}}

		\State $g_{A^\prime} \leftarrow \nabla_{A^\prime} \mathcal{L}(A^\prime ; W, X, y_L)$
        \State $S\leftarrow \nabla_{A^\prime}\odot(-2 A^\prime + 1)$
        \State $e \leftarrow$ top $n$ elements in $S$ if they are not in $R$ 
		\State $A^\prime \leftarrow$ remove or insert edge $e$ from/to $A^\prime$
		\State $R  \leftarrow$ add $e$ in record $R$
\EndWhile  
\State \textbf{return:} $A^\prime$
\end{algorithmic}\label{alg: greedy_attack}
\end{algorithm}
To satisfy the symmetric attribute of $A$, we only perturb the upper triangular part of the matrix $A$ and replicate it to the lower triangular part. This is the reason that in Line 4 we have $2M$. In Line 6, we compute the gradient over $A^\prime$ of Eq.~\eqref{eq:  attack_loss}. In Line 7, we conduct element-wise product between the gradient $g_{A^\prime} $ and a flipped adjacency matrix (0 becomes 1, and 1 becomes -1). This step actually help us express the edge that can be removed or inserted from/to $A^\prime$ also with the gradient value. In Line 8, we speed up the greedy search process. By calculating the $g_{A^\prime} $ once, we select at most $n$ edges to be changed in $A^\prime$, this improvement accelerate the whole attacking method around $n$ times. The sensitivity between the attacking performance and choice of $n$ will also be  shown in experiment part. Line 9 records all the modified edges to avoid sometimes change repeatedly.

Overall, our attack method utilize the gradient information but also satisfies the strict constraints compared to the conventional gradient descent methods. The proposed greedy search step also helps in terms of the convergence speed.

\subsubsection{Zeroth-Order  Greedy  Topology Attack}\label{sec:zo_gta}

Note although during training stage, we usually use sparse matrix to store $A$, when optimizing graph over $A^\prime$,  $g_{A^\prime}$ may be very dense, so that even efficient method like GTA may not be feasible to extremely large graph. Thus, to further improve the efficiency of GTA, we also address the expensiveness of obtaining gradient issue via zeroth-order methods. 
As an extension to our GTA method, we propose  Zeroth-order Greedy Topology Attack (ZO-GTA) to improve the feasibility on large graph. We summarize our ZO-GTA in Algorithm~\ref{alg: zo_greedy_attack}.

\begin{algorithm}
\caption{Zeroth-order Greedy topology attack (ZO-GTA)}
\begin{algorithmic}[1]
\State \textbf{Input:} graph $G = (A,X)$, number of maximum edges can be changed $M$, greedy search step $n$, label $y_L$
\State \textbf{Output:} Modified adversarial adjacency matrix $A^\prime$
\State $A^\prime \leftarrow A$
\State $L^{(0)} \leftarrow$ training loss $ \mathcal{L}(A^\prime; W, X, y_L)$
\State $t \leftarrow 1$
\While{{$\|A^\prime - A\|_0 < 2M$}}
		\State $A^\prime \leftarrow$ random choice $n$ elements in $A^\prime$ and flip them.
        \State $L^t \leftarrow$ training loss $ \mathcal{L}(A^\prime; W, X, y_L)$
        \If{$L^t > L^{t-1} $}
            \State continue
        \Else
            \State $A^\prime \leftarrow$  flip again the $n$ elements in $A^\prime$
        \EndIf
\EndWhile  
\State \textbf{return:} $A^\prime$
\end{algorithmic}\label{alg: zo_greedy_attack}
\end{algorithm}

Intuitively, we continuously check randomly flipped $n$ nodes to see whether they can help us maximize $ \mathcal{L}$ or not. If yes, we keep the $n$ nodes flipped, otherwise, we try this procedure again. The ZO-GTA evade to calculate the gradients in problem~\ref{eq:  attack_loss} and preserve the discrete nature of $A^\prime$ as well.

\subsection{Algorithm for GNN Adversarial Training}

With the aid of our effective attack methods, the robust training for GNNs (see \eqref{eq:  robust_train}) will be solved in this section. 

Note that the inner maximization problem in \eqref{eq:  robust_train} is exactly same as Eq.~\eqref{eq:  attack_loss}. Thus, our GTA methods can approximate this inner optimization. For the outer minimize function, we follow~\cite{kipf2016semi} by using gradient descent to update $W$. This formulation aims to minimize the training loss for the worst case topology perturbations. We summarize our robust training method for GNNs in Algorithm~\ref{alg: adv_training}.

\begin{algorithm}
\caption{Robust training for solving problem \eqref{eq:  robust_train}}
\begin{algorithmic}[1]
\State \textbf{Input:} graph $G = (A,X)$, number of maximum edges can be changed $N$, greedy search step $n$, learning rates $\beta$, and iteration numbers $T$, label $y_L$
\State \textbf{Output:} $W$
\State randomly initialize $W^{(0)}$
\For{$t =  1,2,\ldots, T$}
\State inner minimization over $A^\prime$: given  $ W^{(t-1)}$,   running \hspace*{0.18in}
    GTA~\eqref{eq:  attack_loss} and obtain $A^\prime$
\State outer maximization over $ W$: given  $A^\prime$, obtain  
{\small \begin{align*}
     W^{t} =  W^{t-1} - \beta \nabla_{ W} \mathcal{L}(A^\prime, W^{t-1}; X, y_L)
\end{align*}}%
\EndFor  
\State return $W$
\end{algorithmic}\label{alg: adv_training}
\end{algorithm}
Note that this robust training algorithm is general enough to accommodate any existing attack method as long as constraint $ A^\prime \in \mathcal{C}$ in the inner minimization step is satisfied. For example, one can replace GTA with ZO-GTA or CE-PGD attack~\cite{xu2019topology} in Algorithm~\ref{alg: adv_training} to get different variants of robust models. Different attack methods will results in robust models with different degrees of robustness.

\section{Experiments}
\subsection{Experimental Setup}
This section presents our experimental results for both GTA and robust training on graph convolutional networks (GCN) \cite{kipf2016semi} on three popular datasets: Cora, Citeseer and Pubmed~\cite{sen2008collective}. The datasets statistics are summarized in Table~\ref{table:dataset}.
\vspace{-3mm}
\begin{table}[htb]
 \centering
  \caption{Dataset statistics summary~\cite{yang2016revisiting}} 
  \vspace{1mm}
\begin{tabular}{c|cccccc}
\toprule[1pt]
                 Dataset  & Nodes ($N$) & Edges & Classes & Features  \\
\midrule[1pt]
 Cora    &  2,708   & 5,429   & 7 & 1,433  \\
Citeseer & 3,327    & 4,732   & 6  & 3,703 \\
Pubmed &  19,717  & 44,338  & 3  & 500    \\
 
 \bottomrule[1pt]
\end{tabular}

  \label{table:dataset}
\end{table}
\vspace{-2mm}

We demonstrate the misclassification rate (namely, 1-prediction accuracy on unseen test nodes) of the proposed GTA and ZO-GTA method. 
To provide reliable results, we repeat these experiments $5$ times based on different splits of training/testing nodes and report \textit{mean} $\pm$ \textit{standard deviation} of the performance.
We follow~\cite{zugner2019adversarial,xu2019topology} where for test nodes' predicted labels (not their ground-truth label, generate by an independent pre-trained model) can be used during the attack.

\subsection{Attack Performance}
\begin{figure}[tb]  
\centering
\begin{tabular}{c}
 \includegraphics[width=.4\textwidth]{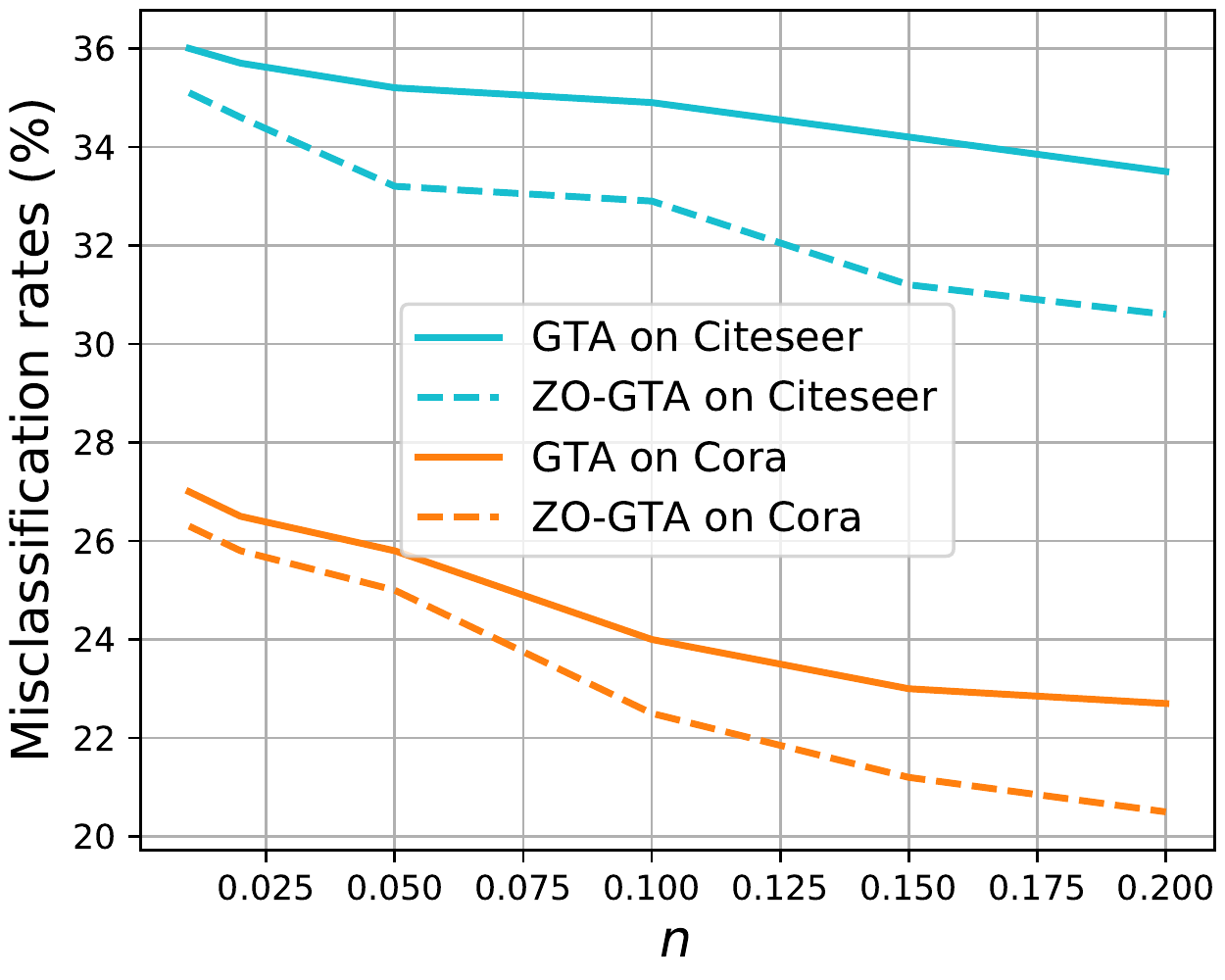}
 \end{tabular}
\caption {Misclassification rate  (in \%) of GTA/ZO-GTA on Citeseer/Cora datasets vs $n$( $n$ is varied between $0.01N$ to $0.2N$).}
\label{fig: ncheck}
\end{figure}

As we mentioned in Section~\ref{sec:attack} and~\ref{sec:zo_gta}, the hyper parameter $n$ affects the speed and the performance of our final model. First we fix $M=0.05 N$ in each dataset.
Next, we conduct the experiment by searching $n \in [0.01N, 0.2N]$ which means we attack a graph by around 5 to 100 iterations. The results are shown in Figure~\ref{fig: ncheck}. As we can see, the misclassification rate drop as $n$ increases. However, the smaller the $n$ is, more iterations we need to execute the attack. Also, we can conclude that GTA performs better than ZO-GTA. 
This is intuitive as ZO-GTA is an approximate scheme as compared to GTA which use gradient information.
Therefore, in the following experiments we set $n = 0.05N$.
We compare our GTA and ZO-GTA methods with DICE (`delete edges internally, connect externally’)~\cite{waniek2018hiding}, CE-PGD and CW-PGD~\cite{xu2019topology}. We follow the hyper parameters and experimental settings as given in in~\cite{xu2019topology} for a fair comparison. The attack results and run time are reported in Table~\ref{table: attack_performance} and Table~\ref{table: attack_time}, respectively.

\begin{table}[htb]
 \centering
  \caption{Misclassification rates (in $\%$) of GTA, ZO-GTA, DICE, CE-PGD and CW-PGD over Cora Citeseer and Pubmed datasets.  No result means that the method is not feasible for this dataset.) 
  } 
  \vspace{1mm}
\begin{tabular}{c|ccc}
\toprule[1pt]
                  & Cora & Citeseer & Pubmed\\
\midrule[1pt]
 Clean      & $18.2\pm 0.1$ & $28.9\pm 0.3$  & $16.9\pm 0.5$   \\
 \midrule
 GTA      & $25.8\pm 0.1$ & $35.2\pm 0.1$  &  $\mathbf{19.3\pm 0.4}$  \\
ZO-GTA    & $24.9\pm 0.3$ & $33.1\pm 0.3$  &  $18.8\pm 0.5$  \\

\midrule

 DICE     & $18.9\pm 0.3$ & $29.8\pm 0.4$ &   $17.4\pm 0.6$  \\
 CE-PGD   & $\mathbf{28.0\pm 0.1}$ & $36.0\pm 0.2$ &-    \\
 CW-PGD   & $27.8\pm 0.4$ & $\mathbf{37.1\pm 0.5}$ &-    \\
 
 \bottomrule[1pt]
\end{tabular}
  \label{table: attack_performance}
\end{table}

It can be clearly seen that GTA and ZO-GTA perform competitively with gradient methods, i.e., CE-PGD and CW-PGD and yield significantly better computational efficiency.

\begin{table}[htb]
 \centering
  \caption{Running time (in seconds) of GTA, ZO-GTA, DICE, CE-PGD and CW-PGD over Cora Citeseer and Pubmed datasets.  No result means that the method is not feasible for this dataset.
  } 
    \vspace{1mm}
\begin{tabular}{c|ccc}
\toprule[1pt]
                  & Cora & Citeseer & Pubmed\\
\midrule[1pt]
 GTA      & $37\pm 1$ & $36\pm 1$  & $109\pm 5$   \\
ZO-GTA    & $\mathbf{19\pm 3}$ & $\mathbf{21\pm 3}$  & $43\pm 5$   \\

\midrule
 CE-PGD   & $147\pm 4$ & $144\pm 4$ &-    \\
 CW-PGD   & $151\pm 5$ & $142\pm 2$ &-    \\
 
 \bottomrule[1pt]
\end{tabular}
  \label{table: attack_time}
\end{table}
\vspace{-4mm}

\subsection{Defense Performance}
We next show the improved robustness of GCN by leveraging our proposed robust training algorithm against different topology attacks.  We set $T = 1000$, $\beta = 0.01$ and $M = 0.05N$.  We evaluate Algorithm~\ref{alg: adv_training} on Cora dataset to compare it with~\cite{xu2019topology}. GTA is used to solve the inner maximization problem since it yields better misclassification rate comapred to ZO-GTA. To test the robustness of the models, we use both GTA and CE-PGD attacking methods to attack them. Results in Table~\ref{table: robust_training} show that the GTA based robust training is competitive with CE-PGD based robust training.
However, recall the results in Table~\ref{table: attack_time}, GTA is much more computationally efficient than CE-PGD which helps our GTA method to handle time sensitive tasks and massive graphs.

\begin{table}[htb]
 \centering
  \caption{Misclassification rates (in $\%$) of two robust training methods against GTA and CE-PGD attack on Cora  dataset. `clean-' means test  without attack.
  } 
    \vspace{1mm}
\begin{tabular}{c|cc}
\toprule[1pt]
                  & GTA training & CE-PGD training \\
\midrule[1pt]
 clean-nature      & \multicolumn{2}{c}{ $18.2\pm 0.1$   } \\
 \midrule
 clean-robust      & $18.1 \pm 0.2$ & $18.0 \pm 0.3$     \\
 \midrule
 GTA attack     & $20.4\pm 0.3$ & $20.8\pm 0.4$     \\
 CE-PGD attack    & $22.7\pm 3$ & $22.0\pm 0.2$   \\

 \bottomrule[1pt]
\end{tabular}
  \label{table: robust_training}
\end{table}

\section{Conclusion}
In this paper, we first introduce two GNN attacking method GTA and ZO-GTA based on the greedy search and zeroth-order algorithms. The proposed approaches are shown to be competitive with state-of-the-art attacks. Next a general and efficient framework of robust training on graph neural networks is proposed. Our experimental results show that the proposed robust training method is significantly computationally less expensive while achieving high robustness to various adversarial attacks. This makes them a potentially viable candidate to handle large-scale and time sensitive problems.

\clearpage

\bibliographystyle{IEEEbib}
\bibliography{refs}

\end{document}